# Collaborative Ensemble Learning: Combining Collaborative and Content-Based Information Filtering via Hierarchical Bayes

Kai Yu,* Anton Schwaighofer,† Volker Tresp
Siemens Corporate Technology, Information and
Communications, 81730 Munich, Germany
kai.yu@gmx.net, volker.tresp@siemens.com,
anton.schwaighofer@gmx.net

Wei-Ying Ma, HongJiang Zhang
Microsoft Research Asia
Beijing, China
wyma@microsoft.com
hjzhang@microsoft.com

## Abstract

Collaborative filtering (CF) and content-based filtering (CBF) have widely been used in information filtering applications, both approaches having their individual strengths and weaknesses. This paper proposes a novel probabilistic framework to unify CF and CBF, named collaborative ensemble learning. Based on content based probabilistic models for each user's preferences (the CBF idea), it combines a society of users' preferences to predict an active user's preferences (the CF idea). While retaining an intuitive explanation, the combination scheme can be interpreted as a hierarchical Bayesian approach in which a common prior distribution is learned from related experiments. It does not require a global training stage and thus can incrementally incorporate new data. We report results based on two data sets, the Reuters-21578 text data set and a data base of user opionions on art images. For both data sets, collaborative ensemble achieved excellent performance in terms of recommendation accuracy. In addition to recommendation engines, collaborative ensemble learning is applicable to problems typically solved via classical hierarchical Bayes, like multisensor fusion and multitask learning.

## 1 INTRODUCTION

The explosive growth of digital media and its usage on the web has opened both great challenges and opportunities to computer science researchers. To handle the increasing amount of available information, several approaches have been developed with the aim of assisting users in finding relevant information. This is often subsumed under the term "information filtering". Examples include:

- *Recommender systems* assist users to find their favorite products like movies, CDs or books, for example on e-commerce web sites

- *Image retrieval*, where the goal is to locate images that match a given query concept (e.g. "images containing red flowers") or user preferences, from large image databases

- *Automatic news filtering* based on news reading habits of a specific user

In the above applications, users typically first provide ratings for a set of exemplar items, (which may be e.g. movies, images or news articles). The information filtering system then returns items that match the interests of the particular user.

Content-based filtering (CBF) and collaborative filtering (CF) represent the two major information filtering technologies. CBF systems analyze the contents of a set of items, together with the ratings provided by an individual user (called the "active user"), to infer which of the yet unseen items might be of interest for the active user. Examples include Balabanovic and Shoham (1997); Mooney and Roy (2000); Pazzani et al. (1996). In contrast, collaborative filtering methods (Resnick et al., 1994; Shardanand and Maes, 1995; Billsus and Pazzani, 1998) typically accumulate a database of item ratings cast by a large set of users. The prediction of ratings for the active user is solely based on the ratings provided by *all* other users. These techniques do not rely on a description of item content. User preferences can be either explicitly expressed by numeric ratings, or implicitly indicated by user behaviors, such as clicking on a hyperlink, purchasing a book or reading a particular news article.

---

*Also with the Institute for Computer Science, University of Munich, Germany

†Also with the Institute for Theoretical Computer Science, Graz University of Technology, Austria

One major difficulty in designing CBF systems lies in the problem of formalizing human perception and preferences based on content analysis. There is a large gap between low-level content features (visual, auditory, or others) and high-level user interests (like or dislike a painting or a CD). Fortunately, the information on personal preferences and interests are all carried in (explicit or implicit) user ratings. Thus CF systems can make use of these high level features rather easily, by combining the ratings of other like-minded users.

On the other hand, pure CF only relies on user preferences, without incorporating the actual content of items. CF often suffers from the extreme sparsity of available data, in the sense that users typically rate only very few items, thus making it difficult to compare the interests of two users. Furthermore, pure CF can not handle items for which no user has previously given a rating. Such cases are easily handled in CBF systems, which can make predictions based on the content of the new item. Therefore, several hybrid approaches have been proposed to compensate the drawbacks of each method.

In this paper, we propose a novel approach to unify CF and CBF in a probabilistic framework, named *collaborative ensemble learning*. Here, a probabilistic model for the preferences of an individual user is built, based on a content description of items (as CBF does). At the prediction phase, collaborative ensemble learning combines the preferences of a society of users, represented by their respective models, to predict an active user's preferences (the CF idea). The combination scheme can be interpreted as a hierarchical Bayes approach in which a common prior distribution is learned from related experiments. Collaborative ensemble learning does not require a global training stage and thus can incrementally incorporate new data. Apart from information filtering, collaborative ensemble learning is also applicable to problems like sensor fusion (Hall and Llina, 2001) and multitask learning (Pratt, 1992; Caruana, 1997; Baxter, 2000; Thrun and O'Sullivan, 1996; Heskes, 2000).

The rest of this paper is organized as follows. After a brief introduction to related work of CBF and CF in Sec. 2, we will describe the idea of modelling user preferences with probabilistic SVMs in Sec. 3. Sec. 4 presents the probabilistic framework to combine a set of user preferences, based on the SVM models, and predict the ratings for the active user. In Sec. 5 we report results for applying collaborative ensemble learning to two data sets for image and text retrieval. We end by giving conclusions and an outlook to future work in Sec. 6.

## 2 RELATED WORK

One of the earliest approaches to content-based information filtering is Rocchio's algorithm (Rocchio, 1971), which implicitly learns the desired content from user feedbacks. Search engines can also be viewed as a kind of content-based system, which filters the web pages or sites based on the contents specified by query words (Pazzani et al., 1996). Mooney and Roy (2000) develop a content-based recommender system using text categorization methods. Recently, content-based image retrieval (CBIR) has also become a vivid research area (Rui et al., 1998).

A variety of CF algorithms have been proposed in the last decade. The earliest memory-based algorithms were based on the observation that people usually trust the recommendations from like-minded friends, such as Resnick et al. (1994) and Shardanand and Maes (1995). Many newly proposed CF methods fall into the class of model-based CF and are inspired from machine learning algorithms. Examples include linear classifiers (Zhang and Iyengar, 2002), Bayesian networks (Breese et al., 1998), dependency networks (Heckerman et al., 2000) and latent class models or mixture models (Hofmann and Puzicha, 1999; Lee, 2001).

Recently many efforts were made to combine collaborative filtering with content-based filtering, mainly based on weighted combinations of CF and CBF systems (Balabanovic and Shoham, 1997; Pazzani, 1999; Basu et al., 1998). There are only few examples of a unifying framework for these two basic information filtering ideas, one being the three-way aspect model of Popescul et al. (2001). Yet, due to the EM-based model fitting and the sparsity of data, this approach may easily suffer from over-fitting.

## 3 MODELLING USER PROFILES

Basis of the proposed collaborative ensemble learning approach are content-based models for the preferences of an individual user $i$. In general, such a preference model can be written as $p(y|x, \theta)$, meaning that it models the distribution of user $i$'s ratings $y$ on some item, described by a vector of features $x$. Collaborative ensemble learning can be used with arbitrary models here, working on discrete or continuous data.

We restrict the discussion here to models for binary preference data. We will present one example for models $p(y|x, \theta)$ that we also use in our experiments, namely a probabilistic version of support vector machines (SVMs).

In the following, we assume a set of $M$ items, each item $j$ being represented by a vector of features $x_j$, $j =$

$1, \ldots, M$. Also, we have preference data for $L$ different users. Preference data for user $i$ consists of a set of rated items $\mathcal{R}_i$, together with a set of ratings $\{y_{ij}\}, j \in \mathcal{R}_i$, where each rating $y_{ij}$ is either $+1$ (liked that particular item) or $-1$ (disliked). The overall preference data for user $i$, $i = 1, \ldots, L$ is denoted by $\mathcal{D}_i = \{(\boldsymbol{x}_j, y_{i,j}) \mid j \in \mathcal{R}_i, \ y_{i,j} \in \{+1, -1\}\}$.

### 3.1 SUPPORT VECTOR MACHINES

Support vector machines (SVMs) are a classification technique with strong backing in statistical learning theory (Vapnik, 1995). They have been applied with great success in many challenging classification problems, including text categorization (Joachims, 1998) and image retrieval (Tong and Chang, 2001).

We consider SVM models for the preferences of user $i$, based on the ratings $\mathcal{D}_i$ this user has previously provided. A standard SVM would predict user $i$'s rating on some item $\boldsymbol{x}$, represented by its feature vector, by computing

$$y = \text{sign}(f^i(\boldsymbol{x})) = \text{sign}\Big( \sum_{j \in \mathcal{R}_i} y_{i,j} \alpha_{i,j} k(\boldsymbol{x}_j, \boldsymbol{x}) + b_i \Big) \quad (1)$$

$k(\cdot, \cdot)$ denotes the kernel function, which computes the pairwise similarities of two items. We will later use $\theta$ to stand for the SVM preference model for user $i$, with $\theta$ containing all SVM model parameters $\alpha_{i,j}$ and $b_i$. The weights $\alpha_{i,j}$ of the SVM are determined by minimizing the cost function

$$C \sum_{j \in \mathcal{R}_i} (1 - y_{i,j} f^i(\boldsymbol{x}_j))_+ + \frac{1}{2} \boldsymbol{\alpha}_i^T K^i \boldsymbol{\alpha}_i \quad (2)$$

By $(\cdot)_+$, we denote a function with $(x)_+ = x$ for positive $x$, and $(x)_+ = 0$ otherwise. $K^i$ is the matrix of all pairwise kernel evaluations on the training data $\mathcal{D}_i$, and $\boldsymbol{\alpha}_i$ is a vector containing all parameters $\alpha_{i,j}$.

### 3.2 PROBABILISTIC EXTENSIONS TO SVM

In their standard formulation, SVMs do not output any measure of confidence for their prediction. Probabilistic extensions of the SVM, where an associated probability of class membership is output, have been independently suggested by several authors. For our work, we follow the idea of Platt (1999), and compute the probability of membership in class $y$, $y \in \{+1, -1\}$ as

$$p(y|\boldsymbol{x}, \theta_i) = \frac{1}{1 + \exp(y A_i f^i(\boldsymbol{x}))} \quad (3)$$

$A_i$ is the parameter[1] to determine the slope of the sigmoid function. This modified SVM retains exactly the same decision boundary $f^i(\boldsymbol{x}) = 0$ as defined in Eq. (1), yet allows an easy approximation of posterior class probabilities. We use a cross validation scheme to set this parameter $A_i$ for each model. Details will be given along with the experimental results in Sec. 5.

So far we have described a model for the preferences of an individual user, based on probabilistic SVMs. Given some training data containing items the user likes and dislikes, this model can predict—based on a description of items using a set of features—an individual user's preferences. SVM models are known for their excellent performance in many challenging classification problems. However, using only the models for individual users would pose the same problems as common CBF methods, in that the models have very high variance (due to the insufficient amount of training data) and only a poor generalization ability. In the following section, we will present a way of combining the individual user models, thus exploiting the knowledge we have gained from possibly like-minded users, to improve the performance of an information filtering system.

## 4  COLLABORATIVE ENSEMBLE LEARNING

Conventional approaches to information retrieval are either solely based on the preferences of other users (collaborative filtering) or solely based on the known preferences of the query user (content based filtering). In this section, we will combine the preferences of the query user, as described by the user's PSVM model, with other users' preferences, again described by their PSVM models and make predictions for the query user. The proposed combination scheme retains an intuitive explanation and has clear links to existing information filtering approaches.

We assume here that we have collected a set of liked and disliked items for each user $i$, denoted by $\mathcal{D}_i$, $i = 1, \ldots, L$. For each user, a PSVM model has been built according to Eq. (1). We summarize the parameters for this model by $\theta$. One of these users is also the query user, where we will use the index $q$ to indicate the corresponding preference (training) data $\mathcal{D}_q$.

---

[1] Platt's original formulation used an additional bias term in the denominator $1 + \exp(y(A_i f^i(\boldsymbol{x}) + b_i))$. Since we typically only have very few training data available, we restrict the model to containing only one additional parameter $A_i$.

### 4.1 A HIERARCHICAL BAYESIAN FRAMEWORK FOR INFORMATION FILTERING

Our modelling assumption is that each user $q$ can be described by a preference model with associated parameter vector $\theta$, and thus makes predictions $p(y|\theta, x)$ where we assume that $y$ has two states $y = +1$ (for "liked") or $y = -1$ (for "disliked"). Here, $x$ describes an actual item to be rated. We assume that items are fixed and given and therefore are not modelled probabilistically. In hierarchical Bayes one now assumes that $\theta$ has been generated as a sample from a prior distribution $p(\theta)$. In contrast to a non-hierarchical framework, one would here assume that $p(\theta)$ is informative in the sense that it can be learned from the ensemble of user profiles. A practical solution to learning $p(\theta)$ in the hierarchical case will be described in the next section.

Under the above assumptions, we can write for the joint distribution of rating $y$ on item $x$ given by user $q$ with profile $\theta$ and preference data $\mathcal{D}_q$

$$p(\theta, \mathcal{D}_q, y|x) = p(\theta)p(\mathcal{D}_q|\theta)p(y|\theta, x)) \qquad (4)$$

where $p(\mathcal{D}_q|\theta) = \prod_k p(y_{q,k}|, x_{q,k}, \theta)$. An equivalent symmetric version of Eq. (4)

$$p(\theta, \mathcal{D}_q, y,|x) = p(\mathcal{D}_q)p(\theta|\mathcal{D}_q)p(y|\theta, x) \qquad (5)$$

goes parallel with our understanding of the information retrieval process: Observing the training data $\mathcal{D}_q$ with prior distribution $p(\mathcal{D}_q)$, the system infers the *a posteriori* distribution of user profiles and then predicts the rating $y$ for some item $x$.

Through a straight-forward application of Bayes's rule and integrating over $p(\theta)$, we get the following expression for the posterior distribution of ratings:

$$p(y|\mathcal{D}_q, x) = \frac{1}{p(\mathcal{D}_q)} E_{p(\theta)}\left[p(\mathcal{D}_q|\theta)p(y|\theta, x)\right] \qquad (6)$$

with $E_{p(\theta)}[\cdot]$ denoting expectation with respect to $p(\theta)$, the prior distribution of user profiles $\theta$, and

$$p(\mathcal{D}_q) = \int p(\mathcal{D}_q|\theta)p(\theta)d\theta = E_{p(\theta)}\left[p(\mathcal{D}_q|\theta)\right] \qquad (7)$$

Typically, the prior distribution is assumed to be uninformative. In contrast, the hierarchical framework of collaborative ensemble learning allows the model to learn the prior from the ensemble of users, as we will describe in the next section.

### 4.2 COLLABORATIVE ENSEMBLE LEARNING

Collaborative ensemble learning can be introduced in a straight-forward way as follows. We assume a given set of profiles of individual users, $\{\theta_1, \ldots, \theta_L\}$, generated from a common prior distribution $p(\theta)$. In general, the user profiles can be modelled in an arbitrary form and one is not restricted to the SVMs used in our work. The goal in the prediction stage is to infer the rating of some query user $q$ on an item $x$. We assume that the preference model $\theta$ of the query user is part of the total set of user profiles (i.e. $q \in [1, \ldots, L]$).

To obtain an estimate of the informative prior distribution we start by integrating out the unknown parameters $\theta$ and obtain the marginal likelihood of all data $\mathcal{D} = \{\mathcal{D}_1, \ldots, \mathcal{D}_L\}$ as

$$p(\mathcal{D}) = \prod_{i=1}^{L} \int p(\theta)p(\mathcal{D}_i|\theta)d\theta. \qquad (8)$$

We can now obtain an improved estimate (in terms of the marginal likelihood) of the prior distribution as[2]

$$\hat{p}(\theta) = \frac{1}{L}\sum_{i=1}^{L} p(\theta|\mathcal{D}_i) \qquad (9)$$

where $p(\theta|\mathcal{D}_i)$ corresponds to the posterior parameter distribution learned for the $i$th user.[3] Substituting this estimate into Eq. (6), we get

$$\hat{p}(y|\mathcal{D}_q, x) = \frac{1}{\hat{p}(\mathcal{D}_q)}\sum_{i=1}^{L} \int p(\mathcal{D}_q|\theta)p(y|\theta, x)p(\theta|D_i)d\theta \qquad (10)$$

The integrals in this solution need to be approximated for most models. We employ the MAP approximation, which leads to our final solution

$$\hat{p}(y|\mathcal{D}_q, x) = \frac{1}{\hat{p}(\mathcal{D}_q)}\sum_{i}^{L} p(\mathcal{D}_q|\theta_i^{\text{MAP}})p(y|\theta_i^{\text{MAP}}, x) \qquad (11)$$

where

$$\hat{p}(\mathcal{D}_q) = \sum_{i=1}^{L} p(\mathcal{D}_q|\theta_i^{\text{MAP}}) \qquad (12)$$

and where $\theta_i^{\text{MAP}}$ is the MAP-estimate of the posterior parameter distribution. Since we assumed an uninformative initial prior we can approximate $\theta_i^{\text{MAP}} \approx \theta_i^{\text{ML}}$ by the maximum likelihood solution $\theta_i^{\text{ML}}$. Alternatively, for binary preference data, the normalization can be obtained by ensuring that $p(y = 1|\mathcal{D}_q, x) + p(y = -1|\mathcal{D}_q, x) = 1$.

---

[2] We just obtain an improved estimate. To maximize the marginal likelihood one would have to iteratively re-calculate $p(\theta|\mathcal{D}_i)$ (E-step) and re-estimate the prior Eq. 9 (M-step). This would be equivalent to full EM learning, which comes at the expense of increasing the complexity of the solution.

[3] For the calculation of $p(\theta|\mathcal{D}_i)$, we assume an uninformative prior distribution for $\theta$.

To obtain the rating for some item $x$, each user's model simply needs to perform a prediction $p(y|\theta_i^{\text{MAP}}, x)$ which subsequently can be absorbed into the weighting terms. In total, the computations required for a new rating are $L$ times the calculation required for an individual user model. If for a given user a sufficient number of new ratings is available, the model for this user can be retrained.

Note that the predicted rating Eq. (11) can be evaluated easily for any kind of models for individual user preferences. In this paper, we will use the probabilistic SVM given in Eq. (3) as the model for an individual user's preferences.

Finally, note that we would also have obtained Eq. 11 by treating each MAP-parameter estimate as a sample from the prior distribution and by approximating the prior using the empirical distribution as $\hat{p}(\theta) = \sum_{i=1}^{L} \delta(\theta - \theta_i^{\text{MAP}})$ where $\delta(\cdot)$ is the delta function.

### 4.3 INTERPRETATION

Eq. (11) can be interpreted as recommendations based on a mixture model with $L$ components. The predicted rating of some given item $x$ under user $i$'s model, $p(y|\theta_i^{\text{MAP}}, x)$ takes on the role of a mixture component. The term $p(\mathcal{D}_q|\theta_i^{\text{MAP}})$ (divided by the normalizing term $\hat{p}(\mathcal{D}_q)$) is the according weight of the component. A higher likelihood of the query user's example data $\mathcal{D}_q$ under some other user $i$'s model indicates that these two persons share similar opinions. Thus, the prediction of the model built for user $i$ should naturally obtain a high weight.

A particular feature of collaborative ensemble learning is that it approaches a content-based method when very much information for a particular query user is available. To see this, note first that we have assumed that the query user $q$'s model is also part of the total set of models resp. profiles. When $\mathcal{D}_q$ becomes large, all terms $p(\mathcal{D}_q|\theta_i^{\text{MAP}})$, $i \neq q$, will be negligible as compared to $p(\mathcal{D}_q|\theta_q^{\text{MAP}})$. Thus, predictions are made solely on the query user's model of preference, as a purely content-based approach would.

## 5 EMPIRICAL STUDY

### 5.1 PSVM PARAMETER TUNING

The PSVM model described in Sec. 3 has a few parameter that need to be set: The SVM models in Eq. (2) require the constant $C$ that gives a weighting of errors on the training data. Furthermore, the kernel function $k(\cdot, \cdot)$ may be parameterized. As the last parameter, we need to tune the slope parameter $A_i$ of the PSVM model in Eq. (3).

In our experiments, we use an SVM the radial basis function kernel for working on the art image retrieval problem, and a linear kernel for text retrieval. The kernel parameters, as well as the constant $C$, are chosen to minimize the leave-one-out error on the training data. Since the training set for most users is very small, this typically leads to overfitting. Thus, the kernel parameters are shared between models, and the optimization is with respect to the average leave-one-out error on all models. For choosing the slope $A_i$ of the sigmoidal function Eq. (3), we follow the three-fold crossvalidation strategy suggested by Platt (1999).

### 5.2 TEXT RETRIEVAL ON REUTERS-21578

First we report results from a controlled simulation based on the Reuters-21578 text data set, a collection of news articles that has been widely used in the research on information retrieval. Each article has been assigned a set of categories (which may be empty). From the total data, we eliminate articles without categories, titles or main texts, and categories with less than 50 articles. The main text for each article was pre-processed by removing stop words, stemming, and calculating TF-IDF (i.e. term frequency inverse document frequency) weights for the stemmed terms. The final data are 36 categories covering a total of $10,034$ articles, where $1,152$ articles belong to more than one category.

For the experiments, we assume that each user is interested in exactly one of these categories (an assumption that has been widely used in the literature on text retrieval). To further simulate the working environment of collaborative ensemble learning, we generate example data for $L = 360$ users by choosing at random a set of $E$ example items. Each example item is labelled either $+1$ (if it falls into the assumed category for this particular user) or $-1$. The example items and their respective labels make up the training data $\mathcal{D}_i$, $i = 1, \ldots, L$. We report results for two scenarios, with $E = 5$ (insufficient information about users) and $E = 30$.

To evaluate collaborative ensemble learning, we examine the average precision on a set of 180 query (test) users where an increasing amount of rated examples available for the query users. This learning curve is evaluated for 2, 5, 10, 20, 50, and 100 known rated examples for the query users. Precision is measured on the top 100 recommended articles, which is the percentage of 100 articles recommended to a user that the user is actually interested in.

In both training and test data, categories have an equal chance to be assigned to users. We draw items (that is, articles) uniformly at random as well, but ensure that at least one positive and one negative example are selected for each user.

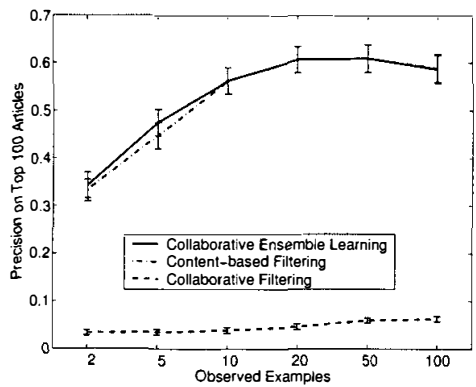

Figure 1: Precision on Reuters data set, averaged over 180 test users, with error bars. For each training user, we use $E = 5$ example articles

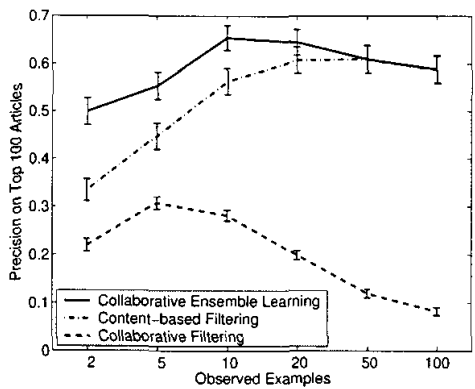

Figure 2: Precision on Reuters data set, averaged over 180 test users, with error bars. For each training user, we use $E = 30$ example articles

Using the experimental setup, we compare collaborative ensemble learning with two other methods for information filtering: (1) collaborative filtering using Pearson correlation (Breese et al., 1998), and (2) content-based filtering using SVM with linear kernel (Drucker et al., 2001; Joachims, 1998).

The results are shown in Fig. 1 (for the case of very little information about each training user, with $E = 5$) and Fig. 2 (for the case $E = 30$). Not surprisingly, collaborative filtering performs worst in both cases. There are two reasons for this point: (1) each training user only visits a very small portion of the corpus (0.05% resp. 0.3%). It is very unlikely that two users have rated the same articles, thus making it very difficult to obtain a reliable estimate of correlation between user interests. (2) A large number of articles is not rated by any training user (90% resp. 70% for the cases $E = 5$ and $E = 30$). Standard collabora-tive filtering can not handle those non-rated articles. Content-based filtering using SVMs provides reasonably good results. This is mainly because that the TF-IDF text features are very effective in representing the topics of articles.

We found that collaborative ensemble learning outperforms both other methods. In the first case, where $E = 5$, the advantage of collaborative ensemble learning over content-based filtering is only small. When query users rate more than 10 articles, both method show almost the same precision. However, in the second case with $E = 30$ (see Fig. 2), collaborative ensemble learning significantly outperforms content-based filtering, especially when only little information is given for the query users. When only two rated examples are given for each query user, collaborative ensemble achieves a precision that is 17% higher than that of content-based filtering. Note that a high precision is particularly important when a user is new to a recommender system, that is, when little information is yet known about the query user.

When more than 50 rated examples are given for each query user, the posterior probability of the query user's own model is getting more prominent in Eq. (11). Thus, collaborative ensemble learning approaches a pure content-based approach. This can be observed in both Fig. 1 and Fig. 2. The final decrease in precision in both figures naturally comes from the fact that fewer positive articles remain in the corpus, thus the achievable precision drops.

### 5.3 ART IMAGE RETRIEVAL

As a second challenging information filtering problem, we evaluated collaborative ensemble learning on an image retrieval problem. Compared to the problem of text retrieval, as presented in Sec. 5.2, image retrieval demonstrates quite different characteristics. The commonly used image features, such as color, shape, and texture, are rather weak indicators of high-level information about an image. In particular, for art images, the user preferences are highly personal. Therefore we expect that collaborative filtering approaches are superior to content-based approaches for art image retrieval.

As the basis of our experiments, we collected user preferences for art images in a web-based survey[4]. Here, users are presented art images, chosen at random out of a total of 642 images, and asked for their opinions (like/dislike/not sure). We so collected data from more than 200 visitors. After removing users who had rated less then 5 items, and users who had rated all of their images with one category (only like resp. only dislike), we retain a total of $L = 190$ users. On average, each of them had rated 89 images. To describe the image content, we used 256 correlogram features, 10 features based on wavelet texture, and 9 features on color moment, giving a 275-dimensional feature vector for each image.

---

[4] The survey can be found on http://honolulu.dbs. informatik.uni-muenchen.de:8080/paintings/index.jsp.

Our experiments use a leave-one-out scheme, in which we pick up one user as the query user and treat all other users as our data base of collected user preferences. This is repeated $L$ times, so that each user becomes query user exactly once. Similar to the setup used in Sec. 5.2, we increase the amount of rated images available for the query users, where we evaluate the performance for 2, 5, 10, 20, 50, and 100 rated images, chosen at random from the collected data for this particular user. This is repeated 10 times with different random choices. We skip a particular setup, if the test user has not rated that many images. The measure of performance is precision in retrieval for the top 20 images recommended to the test user. Note that in many cases this quantity is rather small because we do not know the test user's opinion for all images.

The results are reported in Fig. 3. Content-based filtering, based on the image features only, gives a very poor performance, since the low-level image features are not indicative of the image content. Collaborative filtering performs well, once the number of rated images for the test user is suitably large.

Again, collaborative ensemble learning achieves excellent performance, in particular when very few examples are given for a test user. For example, if only two labelled images are available for the test user, a standard collaborative filtering approach achieves a precision of 4.8%. In contrast, collaborative ensemble reaches a precision of 8.2%, which is higher by a factor of 1.71.[5]

Our results show that collaborative ensemble learning can significantly outperform collaborative filtering when less than 20 examples are known for the test user. However, when the test users have rated more than 50 examples, collaborative filtering achieves the best precision. We do not yet have a clear explanation for this effect. It might result from the following two effect: (1) When many examples are provided, the prediction will be largely influenced by the profile model of the test user, while influences from other users will be relatively suppressed. This seems to work fine for text data, but may be misleading for data like images, where the content can not be described easily by image features alone. (2) On average, each user rated 89 out of 642 images, and each image had on average been rated by 26 users. Thus, the data are rather dense and thus favor collaborative filtering algorithms. In real world applications, where image data bases tend to be huge, one might expect that this advantage of collaborative filtering is less pronounced.

---

[5]Note that we do not know the users' preferences for all images, thus the true precision may still be different from this measured precision.

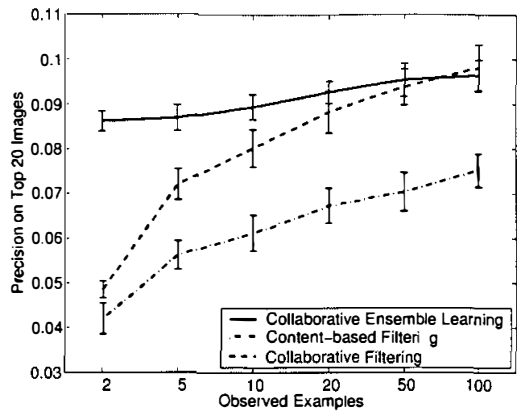

Figure 3: Precision on image retrieval, averaged over 190 test users, with error bars.

Still we can conclude that collaborative ensemble learning achieved an excellent performance. Best performance was achieved when little information is known about a test user—this is the real-world scenario where high performance is most critical.

## 6 EXTENSIONS, RELATED WORK AND CONCLUSIONS

In this paper we presented collaborative ensemble learning, a novel probabilistic solution to combining the basic ideas of content-based and collaborative filtering using a hierarchical Bayesian approach. The resulting equations are very simple, and can be applied to problems where the information between different modules is communicated via a common (hyper-)prior. An example is multisensor data fusion (Hall and Llina, 2001) where a common prior couples the information gained via different sensor channels. Another example is multitask learning in which models trained for different tasks can learn from one another (Pratt, 1992; Caruana, 1997; Baxter, 2000; Thrun and O'Sullivan, 1996; Heskes, 2000). Hierarchical Bayes has a long tradition in statistics and market research and we expect that our solution could be of importance in those fields as well.

We evaluated the performance of the proposed method on two data sets (text and art image retrieval). As compared to pure content-based and collaborative filtering, collaborative ensemble learning achieved excel-lent performance, in particular when very little information is given about the active (query) user. Our experiments have shown that collaborative ensemble learning outperforms its competitors by a large extent in most cases. Since the two data sets show quite different characteristics, one may expect that the promising performance of the method carries over to many other information filtering applications.

In our experiments, we used probabilistic SVMs to model an individual user's (binary-valued) preferences. Collaborative ensemble learning, as described in Sec. 4, can as well be used with other (probabilistic) models for user preferences. Applications to other type of preference data (multinomial, real-valued, and one-class SVMs for user preferences) are currently investigated. In its current form, the number of preference models increases with the number of users. A future extension might thus be to only retain a carefully selected subset of representative user profiles.

## References

Balabanovic M. and Shoham Y. Fab: Content-based, collaborative recommendation. *Communications of the ACM*, 40(3):66–72, 1997.

Basu C., Hirsh H., and Cohen W.W. Recommendation as classification: Using social and content-based information in recommendation. In *Proceedings of the Fifteenth National Conference on Artificial Intelligencen AAAI/IAAI*, pp. 714–720. 1998.

Baxter J. A model of inductive bias learning. *Journal of Artificial Intelligence Research*, 12:149–198, 2000.

Billsus D. and Pazzani M.J. Learning collaborative information filters. In *Proceedings of the 15th International Conference on Machine Learning*, pp. 46–54. Morgan Kaufmann, San Francisco, CA, 1998.

Breese J.S., Heckerman D., and Kadie C. Empirical analysis of predictive algorithms for collaborative filtering. In *Proceedings of the 14th Conference on Uncertainty in Artificial Intelligence*, pp. 43–52. 1998.

Caruana R. Multitask learning. *Machine Learning*, 28(1):41–75, 1997.

Drucker H., Shahrary B., and Gibbon D. Relevance feedback using support vector machines. In *Proceedings of 18th International Conference on Machine Learning*, pp. 122–129. 2001.

Hall D.L. and Llina J., editors. *Handbook of Multisensor Data Fusion*. CRC Press, 2001.

Heckerman D., Chickering D., Meek C., Rounthwaite R., and Kadie C. Dependency networks for inference, collaborative filtering, and data visualization. *Journal of Machine Learning Research*, 1:49–75, 2000.

Heskes T. Empirical bayes for learning to learn. In *Proc. 17th International Conf. on Machine Learning*, pp. 367–374. Morgan Kaufmann, San Francisco, CA, 2000.

Hofmann T. and Puzicha J. Latent class models for collaborative filtering. In *Proceedings of IJCAI'99*, pp. 688–693. 1999.

Joachims T. Text categorization with support vector machines: Learning with many relevant features. In *Proceedings of European Conference on Machine Learning*. Springer, 1998.

Lee W. Collaborative learning for recommender systems. In *Proc. 18th International Conf. on Machine Learning*, pp. 314–321. 2001.

Mooney R. and Roy L. Content-based book recommending using learning for text categorization. In *Proceedings of the Fifth ACM Conference on Digital Libaries*, pp. 195–204. ACM Press, New York, US, San Antonio, US, 2000.

Pazzani M. A framework for collaborative, content-based and demographic filtering. *Artificial Intelligence Review*, 13(5–6):393–408, 1999.

Pazzani M., Muramastsu J., and Billsus D. Syskill and webert: Identifying interesting web sites. In *Proceedings of the Thirteenth National Conference on Artificial Intelligence*, pp. 54–61. Portland, OR, 1996.

Platt J.C. Probabilities for SV machines. In A. Smola, P. Bartlett, B. Scholkopf, and D. Schuurmans, editors, *Advances in Large Margin Classifiers*, pp. 61–74. MIT Press, Cambridge, MA, 1999.

Popescul A., Ungar L., Pennock D., and Lawrence S. Probabilistic models for unified collaborative and content-based recommendation in sparse-data environments. In *17th Conference on Uncertainty in Artificial Intelligence*, pp. 437–444. Seattle, Washington, 2001.

Pratt L.Y. Discriminability-based transfer between neural networks. In *Neural Information Processing Systems (NIPS*5)*, pp. 204–211. 1992.

Resnick P., Iacovou N., Sushak M., Bergstrom P., and Riedl J. Grouplens: An open architecture for collaborative filtering of netnews. In *Proceedings of the 1994 Computer Supported Collaborative Work Conference*, pp. 175–186. ACM, 1994.

Rocchio J.J. Relevance feedback in information retrieval. In *The SMART Retrieval System: Experiments in Automatic Document Processing*, pp. 313–323. Prentice Hall, 1971.

Rui Y., Huang T.S., Ortega M., and Mehrotra S. Relevance feedback: A power tool in interactive content-based image retrieval. *IEEE Trans. Circuits and Systems for Video Tech.*, 8(5):644–655, 1998.

Shardanand U. and Maes P. Social information filtering algorithms for automating 'word of mouth'. In *Proceedings of ACM CHI'95 Conference on Human Factors in Computing Systems*, vol. 1, pp. 210–217. 1995.

Thrun S. and O'Sullivan J. Discovering structure in multiple learning tasks: The tc algorithm. In *Proceedings of the 13th International Conference on Machine Learning*. Morgen Kaufmann, 1996.

Tong S. and Chang E. Support vector machine active learning for image retrieval. In *Proceedings of ACM conference on Multimedia*, pp. 107–118. Ottawa, Canada, 2001.

Vapnik V. *The Nature of Statistical Learning Theory*. Springer, New York, 1995.

Zhang T. and Iyengar V.S. Recommender systems using linear classifiers. *Journal of Machine Learning Research*, 2:313–334, 2002.